\documentclass[conference]{IEEEtran}
\IEEEoverridecommandlockouts
\usepackage{cite}
\usepackage{tikz}
\usepackage{amsmath,amssymb,amsfonts}
\usepackage{algorithmic}
\usepackage{graphicx}
\usepackage{textcomp}
\usepackage{xcolor}
\usepackage[utf8]{inputenc}
\usepackage{newunicodechar}
\usepackage[table,xcdraw]{xcolor} 
\newunicodechar{₁}{$_1$}
\def\BibTeX{{\rm B\kern-.05em{\sc i\kern-.025em b}\kern-.08em
    T\kern-.1667em\lower.7ex\hbox{E}\kern-.125emX}}
\begin{document}

\usetikzlibrary{shapes.geometric, arrows.meta, positioning}

\tikzstyle{startstop} = [rectangle, rounded corners, minimum width=3.5cm, minimum height=1cm,text centered, draw=black, fill=gray!20]
\tikzstyle{process} = [rectangle, minimum width=3.5cm, minimum height=1cm, text centered, draw=black, fill=white]
\tikzstyle{arrow} = [thick,->,>=stealth]

\title{Early Prediction of Multi-Label Care Escalation Triggers in the Intensive Care Unit Using Electronic Health Records\\

}

\author{\IEEEauthorblockN{1\textsuperscript{st} Syed Ahmad Chan Bukhari*}
\IEEEauthorblockA{\textit{Division of Computer Science, Mathematics and Science} \\
\textit{St. John's University}\\
New York, USA \\
Corresponding: bukharis@stjohns.edu}
\and
\IEEEauthorblockN{2\textsuperscript{nd} Amritpal Singh}
\IEEEauthorblockA{\textit{Division of Computer Science, Mathematics and Science} \\
\textit{St. John's University}\\
New York, USA \\
amritpal.singh@stjohns.edu}
\and
\IEEEauthorblockN{3\textsuperscript{rd} Shifath Hossain}
\IEEEauthorblockA{\textit{Division of Computer Science, Mathematics and Science} \\
\textit{St. John's University}\\
New York, USA \\
shifath.hossain20@my.stjohns.edu}
\and
\IEEEauthorblockN{4\textsuperscript{th} Iram Wajahat}
\IEEEauthorblockA{\textit{Institute of Biotechnology} \\
\textit{St. John's University}\\
New York, USA \\
wajahati@stjohns.edu}
}

\maketitle

\begin{abstract}
Intensive Care Unit (ICU) patients frequently exhibit complex, overlapping signs of physiological deterioration that require timely escalations in care. Traditional early warning systems, such as SOFA or MEWS, are limited by their single-outcome focus, failing to capture the multi-dimensional nature of clinical decline. This study proposes a novel multi-label classification framework to predict ``Care Escalation Triggers'' (CETs) precursor states including respiratory failure, hemodynamic instability, renal compromise, and neurological deterioration using the first 24 hours of ICU data.

Using the MIMIC-IV database, CETs are defined through rule-based criteria applied to data from hours 24--72 (e.g., SpO\textsubscript{2} $<$ 90\%, MAP $<$65 mmHg, creatinine increase $>$ 0.3 mg/dL, GCS drop $>$ 2). Features are extracted from the first 24 hours and include vital sign aggregates, key laboratory values, and static demographics. We train and evaluate multiple classification models on a cohort of 85{,}242 ICU stays (using a 80\%, 20\% split - train: 68{,}193; test: 17{,}049). Evaluation metrics include per-label precision, recall, F1-score, and Hamming loss. XGBoost, our best performing model achieves F1-scores of 0.66 (respiratory), 0.72 (hemodynamic), 0.76 (renal), and 0.62 (neurologic), outperforming baseline models. Feature analysis reveals that clinically relevant physiological parameters, such as respiratory rate, blood pressure, and creatinine, are the most influential predictors, aligning closely with the clinical definitions of the respective CETs.

Our framework demonstrates practical potential for early, interpretable clinical alerts without requiring complex time-series modeling or NLP. Its simplicity supports rapid integration into ICU workflows, advancing early warning systems toward multi-faceted, data-driven risk assessments.
\end{abstract}


\section{Introduction}

The Intensive Care Unit (ICU) is a high-stakes environment where critically ill patients require continuous monitoring to detect and manage life-threatening conditions. These patients often exhibit complex, overlapping clinical states-including respiratory failure, hemodynamic instability, renal compromise, or neurological deterioration-that signal the need for escalated care, ranging from intensified monitoring to invasive interventions like mechanical ventilation or vasopressor support. Traditional early warning systems, such as the Sequential Organ Failure Assessment (SOFA) score or the Modified Early Warning Score (MEWS), rely on aggregated risk metrics to predict single outcomes, typically mortality or acute decompensation \cite{b1}, \cite{b2}. While effective for specific use cases, these systems often fail to capture the multifaceted nature of patient deterioration, where multiple organ systems may falter simultaneously, necessitating a more holistic approach to risk assessment.

The advent of comprehensive, time-stamped electronic health record (EHR) datasets, such as the Medical Information Mart for Intensive Care (MIMIC-IV), has revolutionized the application of machine learning (ML) in critical care \cite{b3}, \cite{b4}. These datasets provide granular data on vital signs, comprehensive laboratory measurements, and detailed patient demographics, enabling development of predictive models that leverage the temporal and multivariate nature of ICU data. Recent studies have demonstrated ML's potential to predict single adverse events, such as sepsis, ICU readmission, or mortality, with high accuracy \cite{b5}–\cite{b7}. However, most existing models adopt a single-label classification framework, focusing on one outcome at a time, which fails to capture the fact that patients often deteriorate along multiple physiological dimensions simultaneously. For instance, a patient may simultaneously develop respiratory distress and hemodynamic instability, requiring tailored interventions for each condition.

To address this gap, this study proposes a novel multi-label classification framework to predict a set of ``Care Escalation Triggers'' (CETs)-precursor states that prompt clinicians to intensify monitoring or intervention-over a 48-hour period following ICU admission. We define four CETs-impending respiratory failure, hemodynamic instability, renal compromise, and neurological deterioration-using rule-based thresholds derived from vital signs and laboratory results. By leveraging early data (first 24 hours after admission) from the MIMIC-IV datasets, including aggregated vital signs, laboratory measurements, and static patient data, we train multi-label Logistic Regression, Random Forest, XGBoost and Neural Network models to predict the occurrence of these CETs. This approach prioritizes simplicity, avoiding complex time-series modeling or natural language processing (NLP), to ensure scalability and rapid implementation in resource-constrained ICU settings.

Our work is motivated by the need for proactive, clinician-friendly tools that can flag co-occurring risks early, enabling timely care adjustments before critical interventions are required. Unlike traditional severity scores or single‐outcome models, our multi‐label approach predicts four distinct deterioration pathways-respiratory failure, hemodynamic instability, renal compromise, and neurologic deterioration-over the 48‐hour window following the first day in ICU. This framework not only identifies individual triggers but also quantifies common combinations (for example, respiratory failure co‐occurring with hemodynamic instability), enabling a more holistic view of patient risk. Our multi‐label classifiers achieved strong F₁‐scores for the more prevalent triggers and reasonable performance for the rarer events, outperforming conventional single‐label baselines in recognizing risks. 

\section{Related Work}

The application of machine learning in ICU settings has seen significant growth, driven by the availability of large, publicly accessible datasets like MIMIC-III and MIMIC-IV \cite{b3,b4}. These datasets, comprising de-identified EHR data from Beth Israel Deaconess Medical Center, include time-stamped vital signs, laboratory results, clinical notes, and demographic information for tens of thousands of ICU patients. They have become a cornerstone for developing predictive models in critical care, addressing outcomes such as mortality, length of stay (LOS), readmission, and specific complications like sepsis or acute kidney injury (AKI) \cite{b5,b6,b7,b8,b9,b10}. Below, we review key studies relevant to our multi-label classification approach for predicting care escalation triggers, focusing on ML applications in ICU outcome prediction, multi-label frameworks, and the use of MIMIC data.

\subsection{Machine Learning for ICU Outcome Prediction}

Machine learning has been widely applied to predict single adverse outcomes in the ICU, leveraging the granular data in MIMIC datasets. For instance, Johnson et al. conducted a systematic review of ML applications using MIMIC data, identifying 61 studies that focused on predicting mortality, complications, and readmissions \cite{b5}. Common approaches include Random Forests, Support Vector Machines (SVMs), and deep learning models like Long Short-Term Memory (LSTM) networks, which exploit the temporal nature of EHR data. For example, Kong et al. used logistic regression and gradient boosting to predict in-hospital mortality for sepsis patients, achieving an area under the receiver operating characteristic curve (AUROC) of 0.85 \cite{b6}. Similarly, Yeh et al. applied ML to predict hyperchloremia in critically ill patients, using features like laboratory values and vital signs from MIMIC-III \cite{b7}. These studies demonstrate the power of ML in identifying single high-risk events but often overlook the occurrence of multiple clinical states, limiting their clinical utility in complex ICU scenarios.

ICU readmission prediction is another active area of research, as unplanned readmissions are associated with increased morbidity and mortality. Desautels et al. developed an ensemble model (iREAD) to predict readmission within 48 hours, achieving AUROCs of 0.768 and 0.725 on MIMIC-III and eICU datasets, respectively \cite{b8}. Their model used 30 features, including vital signs and laboratory values, and outperformed traditional scores like the Stability and Workload Index for Transfer (SWIFT) \cite{b9}. Similarly, Lin et al. benchmarked deep learning architectures, including neural ordinary differential equations (ODEs), to predict 30-day readmission risks using MIMIC-III data, achieving comparable performance across architectures \cite{b10}. While these studies highlight the potential of ML for readmission risk, they focus on a single outcome, missing the opportunity to model co-occurring clinical states that may precede readmission.

\subsection{Multi-label Classification in Healthcare}

Multi-label classification, where multiple outcomes are predicted simultaneously, is less common in ICU research but has shown promise in capturing the complexity of patient conditions. Harutyunyan et al. proposed a benchmark suite using MIMIC-III data for four clinical prediction tasks, including phenotype classification, which involved predicting multiple diagnoses (e.g., diabetes, hypertension) from ICU data \cite{b11}. Their multitask learning approach achieved strong performance by jointly training models for all tasks, suggesting that shared representations across outcomes can improve predictive accuracy. Similarly, Sheikhalishahi et al. used a convolutional neural network (CNN) to predict abnormalities in multiple laboratory values (e.g., creatinine, lactate) within a 4-hour window, framing the problem as a multi-label classification task \cite{b12}. Their model, tested on MIMIC-III and eICU data, achieved robust performance by incorporating time-windowed sampling, demonstrating the practicality of multi-label approaches for laboratory-based predictions.

In the context of ICU deterioration, few studies have explicitly targeted multiple care escalation triggers. One notable exception is the work by Ghanbari et al., who developed a deep learning model to predict serum creatinine abnormalities in critically ill patients, incorporating temporal trends from MIMIC-IV \cite{b13}. While their model focused on a single organ system (renal), it demonstrated the value of early data (first 24 hours) in predicting future deterioration, aligning with our approach. However, most multi-label studies in healthcare focus on diagnostic classification (e.g., ICD-9 codes) rather than actionable clinical states like CETs, which are more directly relevant to ICU decision-making.

\subsection{Early Warning Systems and Their Limitations}

Traditional early warning systems like SOFA, MEWS, and the Acute Physiology and Chronic Health Evaluation (APACHE) score are widely used in ICUs to assess patient risk \cite{b1,b2,b14}. These systems aggregate clinical variables into a single score to predict outcomes like mortality or organ failure. For example, the SOFA score, which evaluates six organ systems, is effective for mortality prediction but does not explicitly model co-occurring organ failures as a multi-label problem \cite{b1}. Badawi et al. compared SOFA and MEWS, noting that while both are predictive of adverse events, their single-outcome focus limits their ability to guide nuanced clinical interventions \cite{b2}. ML-based early warning systems have attempted to address these limitations. For instance, Nemati et al. developed a real-time mortality prediction model using MIMIC-III data, achieving an AUROC of 0.85 by incorporating dynamic features like heart rate variability \cite{b15}. However, these models typically predict a single endpoint (e.g., mortality) rather than multiple simultaneous triggers, underscoring the need for our multi-label approach.

\subsection{Gaps and Opportunities}

Despite the progress in ML applications for ICU care, several gaps remain. First, most studies focus on single-label outcomes, neglecting the reality of co-occurring clinical states in ICU patients. Second, while deep learning models like LSTMs and transformers have shown promise in handling time-series data, they often require complex preprocessing and computational resources, limiting their practicality in resource-constrained settings \cite{b10,b11}. Third, existing early warning systems lack the granularity to predict specific, actionable triggers that guide clinical interventions, such as the need for ventilatory support or vasopressors. Our study addresses these gaps by proposing a lightweight, multi-label classification framework that uses readily available tabular data from MIMIC-IV to predict multiple CETs simultaneously. By focusing on early data (first 24 hours) and avoiding complex time-series modeling, our approach ensures scalability and clinical relevance, building on the strengths of prior work while introducing a novel perspective on ICU risk assessment.

\section{Methodology}

We utilized the MIMIC-IV database to develop a lightweight multi-label classification model for predicting Care Escalation Triggers (CETs) within 48 hours following ICU admission (i.e., hours 24–72 post-\textit{intime}). The cohort included all adult ICU admissions (age $\geq$ 18) with recorded charted vital signs and laboratory values. Only the first ICU stay per hospital admission was considered, yielding a total of 85,242 unique ICU stays.

From each stay, we extracted demographic and clinical features from the first 24 hours of ICU admission. These included:
\begin{itemize}
    \item Demographics: age and gender
    \item Vital signs: mean, minimum, and maximum values of oxygen saturation (SpO$_2$), systolic blood pressure (SBP), mean arterial pressure (MAP), heart rate (HR), and respiratory rate (RR)
    \item Laboratory values: most recent value of creatinine       \\ (lactate, and pH were considered as well, however were dropped due to the labs not being present for a large number of stays - 39,401 and 37,841 respectively)
\end{itemize}
 
\begin{figure}
    \centering
    \includegraphics[width=1\linewidth]{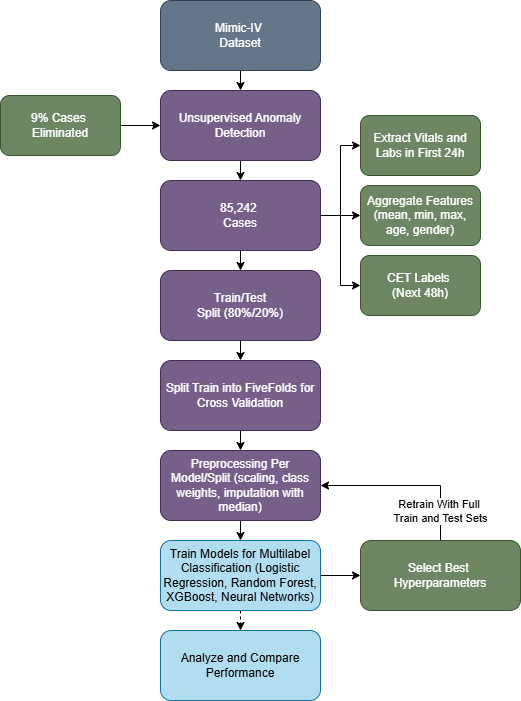}
    \caption{Flow Diagram of Early Prediction of Multi-Label Care Escalation
Triggers in ICU}
    \label{fig:enter-label}
\end{figure}

While imputation techniques could theoretically address certain missing laboratory values, such methods were deemed infeasible given the substantial proportion of absent data (approximately 46\% for lactate and 44\% for pH across the cohort), which could introduce significant bias or reduce model reliability in a machine learning context. Inclusion of lactate and pH, if more available, would likely enhance predictive performance, particularly for respiratory failure and hemodynamic instability CETs, by capturing metabolic and acid-base imbalances that align with their clinical definitions. Age was computed manually by subtracting an individuals anchor age from an estimated date of birth (July 1st of their anchor year). Then the age was calculated as the time difference between the admission date and this approximate date of birth, expressed in years. July 1st was chosen as a mid-year proxy to minimize bias from the unknown exact birthdate, providing a consistent and neutral reference point for age estimation. Gender was encoded using one-hot encoding.

We defined four binary CET outcomes based on:
\begin{itemize}
    \item \textbf{Respiratory Failure:} Two or more SpO$_2$ readings $<$ 90\% or RR $>$ 30 bpm
    \item \textbf{Hemodynamic Instability:} Any MAP $<$ 65 mmHg or SBP $<$ 90 mmHg 
    \item \textbf{Renal Compromise:} Increase in serum creatinine $>$ 0.3 mg/dL and peak creatinine $>$ 1.2 mg/dL
    \item \textbf{Neurologic Deterioration:} Drop in Glasgow Coma Scale (GCS) score $>$ 2 points or any sedation monitoring event
\end{itemize}

The dataset was initially partitioned into training (68,193 samples) and testing (17,049 samples) subsets using an 80\%/20\% split. MultilabelStratifiedShuffleSplit (from the iterstrat library) ensured the proportional representation of multi-label care-escalation triggers across both sets, and MultilabelStratifiedKFold further divided the training data into five evenly distributed folds for cross-validation. Four baseline classifiers -Logistic Regression, Random Forest, XGBoost, and Neural Networks- were implemented. To effectively capture label co-occurrence, a Label Powerset Transformation was applied, converting the multi-label problem into a multi-class one with 16 unique label combinations (derived from our 4 original classes). Hyperparameter tuning was performed using Grid Search Cross-Validation. 
Imputation of the median and weighting inversely proportional to class frequencies were applied to each fold to optimize performance and prevent data leakage. Scaling was also applied on all, but tree-based models (RandomForest and XGBoost) as splits occur on individual features.   For each model, hyperparameter selection was performed by identifying the combination of parameters that maximized the mean macro-averaged F1-score, across all four labels and all cross-validation folds. These metrics were calculated by decoding the Label Powerset predictions back into their original multi-label format, enabling the computation of individual label-specific F1-scores which were then macro-averaged. 

The final models were retrained on the entire training set with the identified optimal hyperparameters. The final hyperparameter configurations were: 
\begin{itemize}
    \item 

    \textbf{Logistic Regression:} "C": 100
    \item \textbf{Random Forest:}"n-estimators": 200, "max-depth": 12,"min-samples-split": 2, "min-samples-leaf": 1
    \item \textbf{XGBoost:} "n-estimators": 200, "max depth": 8, "learning rate": 0.1, "subsample": 0.8
    \item \textbf{Neural Networks:} "hidden-dim": 128, "lr": 0.001, "batch-size": 32, "epochs": 10, "num-hidden-layers": 2
\end{itemize}
The same preprocessing measures (median imputation, feature scaling for non-tree models, and inverse class frequency weighting) were applied to the entire training set prior to final model training. For evaluation, predictions were made on the held-out test set using the 16-class Label Powerset encoding, and these predictions were then decoded back to the original four individual labels to assess model performance. We used standard evaluation metrics for each CET outcome: precision, recall, F1-score, and area under the receiver operating characteristic curve (ROC-AUC). To further analyze the results and understand feature contributions, permutation importance was employed, specifically calculated for each of the four original CETs.

\section{Results}

\begin{table}[htbp]
\caption{Macro Performance Metrics and Top Predictors for CET Labels}
\centering
\footnotesize
\setlength{\tabcolsep}{4pt} 
\renewcommand{\arraystretch}{1.1} 
\rowcolors{2}{gray!10}{white} 
\begin{tabular}{|l|r|r|r|r|}
\hline
\rowcolor{blue!60}
\textcolor{white}{\textbf{Model/CET Label}} & 
\textcolor{white}{\textbf{Accuracy}} & 
\textcolor{white}{\textbf{Precision}} & 
\textcolor{white}{\textbf{Recall}} & 
\textcolor{white}{\textbf{F1}} \\
\hline
\rowcolor{blue!30}
\textbf{Logistic Regression} &  &  &  &  \\
\hline
Respiratory Failure & 0.66 & 0.60 & 0.66 & 0.59 \\
\hline
Hemodynamic Instability & 0.63 & 0.63 & 0.63 & 0.62 \\
\hline
Renal Compromise & 0.76 & 0.65 & 0.74 & 0.66 \\
\hline
Neurologic Deterioration & 0.52 & 0.52 & 0.53 & 0.50 \\
\hline
\rowcolor{blue!30}
\textbf{Random Forest} &  &  &  &  \\
\hline
Respiratory Failure & 0.76 & 0.64 & 0.68 & 0.65 \\
\hline
Hemodynamic Instability & 0.71 & 0.70 & 0.71 & 0.70 \\
\hline
Renal Compromise & 0.83 & 0.72 & 0.82 & 0.75 \\
\hline
Neurologic Deterioration & 0.68 & 0.57 & 0.56 & 0.56 \\
\hline
\rowcolor{blue!30}
\textbf{XGBoost} &  &  &  &  \\
\hline
Respiratory Failure & 0.77 & 0.65 & 0.68 & 0.66 \\
\hline
Hemodynamic Instability & 0.73 & 0.72 & 0.72 & 0.72 \\
\hline
Renal Compromise & 0.85 & 0.74 & 0.81 & 0.76 \\
\hline
Neurologic Deterioration & 0.71 & 0.63 & 0.62 & 0.62 \\
\hline
\rowcolor{blue!30}
\textbf{Neural Networks} &  &  &  &  \\
\hline
Respiratory Failure & 0.67 & 0.60 & 0.66 & 0.59 \\
\hline
Hemodynamic Instability & 0.67 & 0.66 & 0.67 & 0.66 \\
\hline
Renal Compromise & 0.84 & 0.73 & 0.83 & 0.76 \\
\hline
Neurologic Deterioration & 0.66 & 0.59 & 0.59 & 0.59 \\
\hline
\end{tabular}
\label{tab:cet_performance}
\end{table}

On the held-out test set (n=17,049), XGBoost demonstrated superior performance among the tested models, consistently outperforming Logistic Regression. It also closely matched the strong performance of Random Forests and Neural Networks across key evaluation metrics. As detailed in Table 1, the average F1-scores on a per-model basis across all CETs were 0.69 for XGBoost, 0.67 for Random Forest, 0.65 for Neural Networks, and 0.59 for Logistic Regression, clearly illustrating the leading performance of tree-based models (XGBoost and Random Forest), with Neural Networks performing strongly just behind them, and the comparatively lowest performance of Logistic Regression. Figure 2 further reinforces this trend, showing XGBoost and Random Forest consistently achieved the highest ROC-AUC scores (average AUCs of 0.80 and 0.76, respectively), with Neural Networks in the middle (average AUC 0.75), and Logistic Regression again demonstrating the lowest performance (average AUC 0.71).

\begin{figure}
    \centering
    \includegraphics[width=1\linewidth]{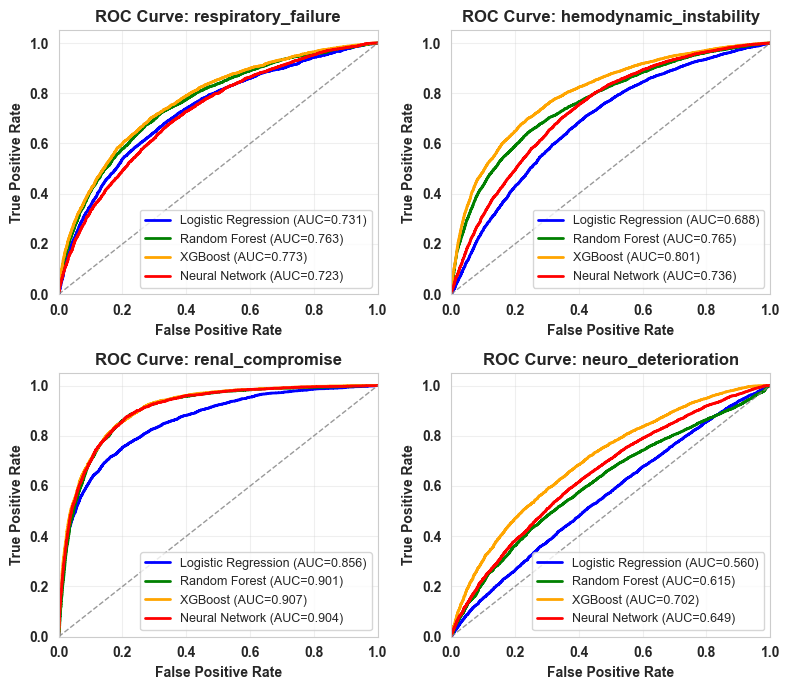}
    \caption{Receiver Operating Characteristic (ROC) Curves for Predictive Models Across Different Care Escalation Triggers}
    \label{fig:enter-label}
\end{figure}

This observed pattern of performance can be attributed to the inherent characteristics of the models and the complexity of the dataset. Logistic Regression likely exhibited the lowest performance due to its linear nature, which may struggle to capture the intricate, non-linear relationships and interactions present in data where multiple Care Escalation Triggers can occur simultaneously.
Conversely, XGBoost and Random Forest consistently demonstrated superior performance, likely benefiting from their ensemble learning approaches. These methods are well-suited for handling high-dimensional, complex datasets with intricate feature interactions, as they can effectively model non-linear relationships and are robust to outliers and noisy data. Neural Networks occupied an intermediate performance position. Despite their capacity for learning complex representations, the performance of Neural Networks is highly sensitive to the initial data representation and the suitability of the chosen architecture for the specific feature space. Tree-based methods naturally excel at handling various feature types and discovering complex interactions within tabular data. In contrast, neural networks may necessitate more deliberate feature engineering or purpose-built architectural configurations to effectively leverage all available information and achieve superior performance.

\begin{figure}
    \centering
    \includegraphics[width=1\linewidth]{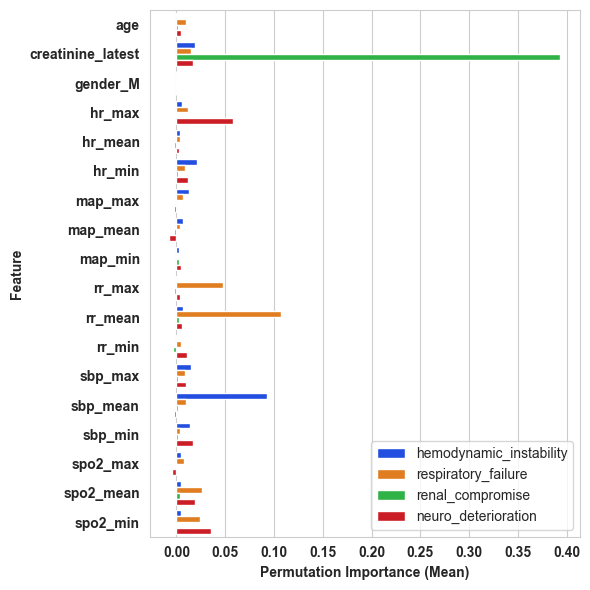}
    \caption{Permutation Importances of XGBoost Model by Care Escalation Triggers}
    \label{fig:enter-label}
\end{figure}

To understand the contributions of input variables to our best-performing XGBoost model's predictions, we performed a permutation importance analysis. This was conducted specifically for each of the four original care-escalation triggers, and the top three predictive features for each type are presented below with their corresponding mean importances:
\begin{itemize}
    \item \textbf{Respiratory Failure:} Mean Respiratory Rate (0.107), Maximum Respiratory Rate (0.048), Mean Blood Oxygen Saturation (0.026), 
    \item \textbf{Hemodynamic Instability:} Mean Systolic Blood Pressure (0.093), Minimum Heart Rate (0.022), Latest Creatinine Level (0.019)
    \item \textbf{Renal Compromise:} Most Recent Creatinine Level (0.394), Mean Blood Oxygen Saturation (0.003), Minimum Mean Arterial Pressure (0.002)
    \item \textbf{Neurological Deterioration:} Maximum Heart Rate (0.058), Minimum Blood Oxygen Saturation (0.035), Mean Blood Oxygen Saturation (0.019)
\end{itemize}
The permutation importance analysis of our XGBoost model offers crucial insights into how early ICU data contributes to predicting specific care escalation triggers (CETs). Unlike traditional feature importance, which can be biased, permutation importance reliably quantifies a feature's predictive contribution by measuring the drop in model performance when its values are randomly shuffled. For Respiratory Failure, the top features -mean and max respiratory rate, and mean blood oxygen saturation- align well with the clinical definitions of this CET (e.g., respiratory rate $>$ 30 bpm or SpO2 $<$ 90\%). This reinforces that early trends in respiratory rate and oxygen saturation are strong indicators of future respiratory deterioration. Similarly, Hemodynamic Instability was primarily driven by mean systolic blood pressure, which is clinically intuitive given the role of hypotension (SBP $<$ 90 mmHg or MAP $<$ 65 mmHg) in its definition. The prominence of heart rate metrics also suggests the model effectively captures cardiovascular compensation.

Renal Compromise showed the most striking pattern: "Most Recent Creatinine Level" had a very high permutation importance ($\sim$0.394). This directly reflects the CET's definition, which relies on changes in serum creatinine, and supports clinical understanding of acute kidney injury (AKI) indicators \cite{b5}. In contrast, feature importance for Neurological Deterioration was more diffuse. Since our model lacked direct Glasgow Coma Scale (GCS), it likely inferred risk from indirect markers like heart rate and minimum blood oxygen saturation. This broader spread suggests that predicting neurological decline from early physiologic trends alone  might be less straightforward, highlighting a potential need for richer or more specific proxy features. Overall, this analysis not only corroborated our clinical definitions of CETs but also showcased the model's interpretability, as features central to defining outcomes were consistently among the most predictive.

\section{Limitations and Future Work}
Our feature extraction relies on aggregated statistics (mean, minimum, and maximum) from the first 24 hours of ICU data, which simplifies processing but may overlook nuanced temporal trends, such as the slope of creatinine changes or variability in vital signs over time. This aggregation prioritizes computational simplicity and real-time applicability but potentially limits the model's ability to detect time-sensitive deterioration patterns. Additionally, the reliance on MIMIC-IV, derived from a single U.S. academic medical center (Beth Israel Deaconess Medical Center), introduces potential biases related to institutional practices, patient demographics, and EHR documentation. Furthermore, neurologic assessments such as Glasgow Coma Scale scores were frequently missing or inconsistently recorded, often confounded by sedation or
intubation, which may have reduced the reliability of neurologic features.These factors may limit generalizability to diverse ICU populations, such as those in non-U.S. settings or smaller hospitals with varying resource levels and EHR systems. External validation on multi-center datasets, such as eICU, would strengthen claims of broader applicability and address these transferability limitations.

Future extensions could incorporate lightweight time-aware methods to enhance performance without requiring complex architectures. For instance, sliding windows could segment the 24-hour data into overlapping intervals, allowing the model to capture short-term fluctuations in vital signs. Alternatively, recurrent models like Long Short-Term Memory (LSTM) networks could process sequential data directly, modeling dependencies across time steps to better predict co-occurring CETs. Regarding deployment, our framework leverages lightweight, off-the-shelf models like XGBoost, which exhibit low computational overhead suitable for real-time ICU integration (e.g., inference times typically under 1 second per prediction on standard hardware, with minimal memory usage). Future work could include pilot deployments, such as integrating the model with EHR alert systems for proactive clinician notifications, to evaluate latency in clinical environments.

\section{Conclusion}

This study successfully demonstrates the feasibility of developing a real-time, multi-label early warning system for predicting Care Escalation Triggers (CETs) using early ICU data. Specifically, Random Forest and XGBoost classifiers, trained on features extracted from the first 24 hours of ICU stay, consistently demonstrated strong performance in identifying patients at risk of all four CETs-respiratory failure, hemodynamic instability, renal compromise, and neurologic deterioration-within the subsequent 48 hours.

The predictive strength of readily available and clinically intuitive features like respiratory rate, oxygen saturation, blood pressure, and creatinine further underscores the clinical relevance and interpretability of our approach. These results strongly support the implementation of machine learning-based early warning systems leveraging readily available electronic health record data to enhance proactive care and mitigate critical deterioration in ICU patients.

\section*{Acknowledgment}
This work is supported by the National Science Foundation under Award No. 2431840. The authors gratefully acknowledge the PhysioNet team for providing access to the MIMIC-IV database, which made this work possible.

\vspace{12pt}


\begin{thebibliography}{00}

\bibitem{b1} J. L. Vincent et al., ``The SOFA (Sepsis-related Organ Failure Assessment) score to describe organ dysfunction/failure,'' \textit{Intensive Care Med.}, vol. 22, no. 7, pp. 707--710, Jul. 1996, doi: 10.1007/BF01709751.

\bibitem{b2} O. Badawi and M. J. Breslow, ``Readmissions and death after ICU discharge: Development and validation of two predictive models,'' \textit{Crit. Care Med.}, vol. 40, no. 11, pp. 2957--2963, Nov. 2012, doi: 10.1097/CCM.0b013e31825fd6e5.

\bibitem{b3} A. E. W. Johnson et al., ``MIMIC-III, a freely accessible critical care database,'' \textit{Sci. Data}, vol. 3, no. 160035, May 2016, doi: 10.1038/sdata.2016.35.

\bibitem{b4} A. E. W. Johnson et al., ``MIMIC-IV, a publicly available ICU dataset,'' \textit{Sci. Data}, vol. 7, no. 1, pp. 1--8, Apr. 2020, doi: 10.1038/s41597-020-0440-9.

\bibitem{b5} A. E. W. Johnson et al., ``Application of machine learning in intensive care unit (ICU) settings using MIMIC dataset: Systematic review,'' \textit{Informatics}, vol. 8, no. 1, p. 16, Mar. 2021, doi: 10.3390/informatics8010016.

\bibitem{b6} G. Kong, K. Lin, and Y. Hu, ``Using machine learning methods to predict in-hospital mortality of sepsis patients in the ICU,'' \textit{BMC Med. Inform. Decis. Mak.}, vol. 20, no. 1, p. 251, Oct. 2020, doi: 10.1186/s12911-020-01271-0.

\bibitem{b7} P. Yeh, Y. Pan, L. N. Sanchez-Pinto, and Y. Luo, ``Using machine learning to predict hyperchloremia in critically ill patients,'' in \textit{Proc. IEEE Int. Conf. Bioinformatics Biomed.}, Nov. 2019, pp. 1703--1707, doi: 10.1109/BIBM47256.2019.8982933.

\bibitem{b8} T. Desautels et al., ``Multicenter validation of a machine learning model to predict intensive care unit readmission within 48 hours after discharge,'' \textit{J. Crit. Care}, vol. 65, pp. 1--7, Oct. 2021, doi: 10.1016/j.jcrc.2021.05.001.

\bibitem{b9} A. Gaj et al., ``Predicting unplanned 7-day intensive care unit readmissions with machine learning models for improved discharge risk assessment,'' \textit{J. Med. Syst.}, vol. 44, no. 9, p. 156, Aug. 2020, doi: 10.1007/s10916-020-01620-7.

\bibitem{b10} Y. W. Lin et al., ``Benchmarking deep learning architectures for predicting readmission to the ICU and describing patients-at-risk,'' \textit{Sci. Rep.}, vol. 10, no. 1, p. 1559, Jan. 2020, doi: 10.1038/s41598-020-58374-8.

\bibitem{b11} H. Harutyunyan et al., ``Multitask learning and benchmarking with clinical time series data,'' \textit{Sci. Data}, vol. 6, no. 96, Jun. 2019, doi: 10.1038/s41597-019-0103-9.

\bibitem{b12} S. Sheikhalishahi et al., ``Predicting abnormalities in laboratory values of patients in the intensive care unit using different deep learning approaches,'' \textit{JMIR Med. Inform.}, vol. 8, no. 7, p. e19127, Jul. 2020, doi: 10.2196/19127.

\bibitem{b13} G. Ghanbari et al., ``Development and validation of a deep learning algorithm for the prediction of serum creatinine in critically ill patients,'' in \textit{Proc. IEEE Int. Conf. Bioinformatics Biomed.}, Nov. 2019, pp. 1708--1712, doi: 10.1109/BIBM47256.2019.8982934.

\bibitem{b14} W. A. Knaus et al., ``The APACHE III prognostic system: Risk prediction of hospital mortality for critically ill hospitalized adults,'' \textit{Chest}, vol. 100, no. 6, pp. 1619--1636, Dec. 1991, doi: 10.1378/chest.100.6.1619.

\bibitem{b15} S. Nemati et al., ``An interpretable machine learning model for accurate prediction of sepsis in the ICU,'' \textit{Crit. Care Med.}, vol. 46, no. 4, pp. 547--553, Apr. 2018, doi: 10.1097/CCM.0000000000002936.

\end{thebibliography}
\end{document}